\documentclass[10pt,twocolumn,letterpaper]{article}
\usepackage{icb}
\usepackage{times}
\usepackage{epsfig}
\usepackage{graphicx}
\usepackage{amsmath}
\usepackage{amssymb}
\usepackage[linesnumbered, ruled]{algorithm2e}
\SetKwRepeat{Do}{do}{while}%
\usepackage{float}
\usepackage{epstopdf}
\makeatletter
\def\BState{\State\hskip-\ALG@thistlm}
\makeatother
\icbfinalcopy % *** Uncomment this line for the final submission

\ificbfinal\fi
\begin{document}

%%%%%%%%% TITLE
\title{Feature Level Fusion from Facial Attributes for Face Recognition}
\author{Mohammad Rasool Izadi\\
University of Notre Dame\\
{\tt\small mizadi@nd.edu}}

\maketitle
\thispagestyle{empty}

%%%%%%%%% ABSTRACT
\begin{abstract}
   We introduce a deep convolutional neural networks (CNN) architecture to classify facial attributes and recognize face images simultaneously via a shared learning paradigm to improve  the accuracy for facial attribute prediction and face recognition performance. In this method, we use facial attributes as an auxiliary source of information to assist CNN features extracted from the face images to improve the face recognition performance. Specifically, we use a shared CNN architecture that jointly predicts facial attributes and recognize face images simultaneously via a shared learning parameters, and then we use facial attribute features an an auxiliary source of information concatenated by face features to increase the discrimination of the CNN for face recognition. This process assists the CNN classifier to better recognize face images. The experimental results show that our model increases both the face recognition  and facial attribute prediction performance, especially for  the identity attributes such as gender and race. We evaluated our method on several standard  datasets labeled by identities and face attributes and the results show that the proposed method outperforms state-of-the-art face recognition models.
\end{abstract}

%%%%%%%%% BODY TEXT
\section{Introduction}
Convolutional Neural Networks (CNNs), have made important enhancement in biometric and computer vision applications  such as face image retrieval \cite{Cao_THN_2017,talreja2018using,taherkhani2018facial,Yang_2017_Pairwise} , attribute prediction \cite{talreja2019attribute,taherkhani2018deep} and  machine learning applications such as Generative Adverse rial Network (GAN) \cite{goodfellow2014generative,kazemi2018unsupervised2} and semi-supervised learning \cite{izadi2020optimization,mohsenvand2020contrastive,iscen2019label,taherkhani2019matrix,liu2019exploiting}. In spite of this improvement, creating a deep architecture to learn several tasks simultaneously while enhancing their accuracy by sharing the learning parameters is an ongoing problem. Providing privileged data (i.e. auxiliary data) such as facial attributes to CNN based architecture to recognize face images is  a costly or tedious task; or the privileged data may not be available during the testing step. Despite the potential advantages of using auxiliary data, these problems have diminished the popularity and ease of use of multi-modality models in face recognition.

In this study, we use facial attributes as a soft modality to enhance face recognition.  We also indicate that when a deep CNN model is trained to recognize face images, the model indirectly learns the facial attributes as well. Specifically, the two modalities enhance the overall prediction performance when they are trained simultaneously. We show that some auxiliary information provided from identity facial attributes  such as gender, nose shape, lips shapes, race can be incorporated with each other and then alongside the facial features, they can assist the CNN to better perform face recognition task. Moreover, it has been shown that variations to illumination, noise, occlusion, and pose confound the identification of a face. However,  some facial attributes, stand as more invariant semantic features from which face recognition, verification and many other applications such as zero-shot learning and user annoyance prediction can benefit \cite{taherkhani2017restoring, akata2013label,christie2014predicting}.

In spites of significant enhancements in face verification task, it is still a challenging task in biometrics. Some methods which use facial attributes are considerable \cite{gong2013hidden}; to cancel out aging variation of face images, Gong et al \cite{gong2013hidden} introduce  an age invariant face recognition algorithm; this method leverages a probabilistic model with two latent factors to separate aging variations from person specific features.  Wen et al \cite{wen2016latent} use a latent identity factor  to guide  CNN learning  parameters; the model extracts  age invariant deep face features  by  learning  latent identity and CNN parameters jointly. Yandong et al \cite{wen2016discriminative} propose a deep CNN based model with a trainable and smart supervision signal called center loss which finds a center for deep features of each class. The center loss penalizes the feature if it is far from its corresponding class center. The method trains softmax and center losses jointly to dispense inter-class samples and compact intra-class  samples.
 
Facial attributes are invariant and robust visual features that can  be recognized  from the face images straightly \cite{torfason2016face}. Attribute recognition algorithm are essentially categorized to local and global methods. Local methods include three stages; in the first step, these methods detect different parts of an object, in the second step, they extract features from each part. In the third step, these features are concatenated for training  a classifier \cite{kumar2009attribute,bourdev2011describing, chung2012deep,berg2013poof,luo2013deep,zhang2014panda}. For instance, The work presented in \cite{kumar2009attribute} performs based on extracting hand-crafted
features from ten parts of the parts. Zhang et al \cite{zhang2014panda} extract poselets which aligns human body parts to recognize human body attributes.  This algorithm works inefficiently if object localization and alignment is not performed properly. Moreover, local methods  do not perform well  if there exist unconstrained face images with complex variations
  in the image. This is because, this factor causes face localization and alignment challenging. Global methods, however, extract features from entire image part of the image ignoring object parts. It is worth mentioning that attribute prediction has been also enhanced in the recent years. For example, Bourdev et al \cite{bourdev2016pose} introduce a part-based attribute prediction algorithm deploying semantic segmentation to transfer localization information to the facial attribute prediction task. In the other case,  Liu et al \cite{liu2015deep} employs two cascaded CNNs; the first one, is used for face localization, while the second one, ANet, is used for attribute description. Moreover, Zhong et al \cite{zhong2016face} first localize images and then use an off-the-shelf architecture created for face recognition to introduce the facial attributes at different levels of a CNN model. He et al \cite{hemulti} introduce a multi-task architecture to predict relative facial attributes. This model applies a CNN model to learn local context and global style information from the intermediate convolution layers and fully connected layers, accordingly. 
 
 In this work, we propose a model guiding CNN parameters to learn face attributes while simultaneously being trained on the objective of face recognition.  We show jointly learning certain facial attributes increases the network's face recognition accuracy. Moreover, we observe that the resultant jointly-trained network is a more capable face attribute classifier than one trained on face attributes alone. The framework is constructed from two cascaded networks which communicate information together by sharing their learning parameters in the model. Facial attributes are extra information which compact samples of a class in a discriminate feature space if they are used to describe input data.  For example, assume a scenario where it is difficult to recognize the individuals in two face images. If a classifier is trained to utilize the fact that one individual is a male and the other is female, the dimension of gender in the input data becomes a potentially discriminating feature. The parameters shared in the model are trained by modified AdaMax optimizer. The rest of this paper is organized as follows: we describe our model in the ‘Proposed model’ section, how we train the model parameters using the modified AdaMax optimizer in 'Training model', and finally results and concluding remarks in ‘Experiments’ and ‘Conclusion’.
%\begin{figure*}[ht]
%\includegraphics[scale=0.26]{Architecture.png}
%\caption{Proposed model, face recognition and attribute prediction are trained simultaneously. }
%%\label{fig:arc}
%\end{figure*}

\section{Facial Attributes for Face Recognition}

The proposed architecture predicts attributes and uses them as an auxiliary modality to recognize face images. The model is created from two networks. The first network uses the VGG structure with same filters size, convolutional layers and pooling operation. The first network uses filters with receptive field  $3 \times 3$.  The convolution stride is 1 pixel. Spatial padding of the convolutional layer is set to 1 pixel for all $3 \times 3$ convolutional layers to keep spatial resolution after convolution operation. Spatial pooling is conducted by five max-pooling layers. Max-pooling is performed on a $2 \times 2$ pixel window with stride 2. Hidden layers are followed by rectification ReLU \cite{krizhevsky2012imagenet} non-linearity operation. 

The second network is split into two separate branches trained jointly while communicating information together via the training stage. Both of these two branches include two fully connected layers followed by the first network. The first FC layers of the branches has 4096 channels. The last layer is a soft-max layer. The first branch conducts the task of attribute prediction  and communicates information by concatenating the attribute soft-max layers with last pooling layer of first network. This information is used to train the second branch which is the face recognition task.
\section{Training the Architecture} 

In this section, we explain how we train our architecture.  Thousands number of images are required to train  a deep model like VGG. For sake of this, we fine tune our architecture using a pre-trained model; we train the first network in the model from scratch using CASIA dataset. CASSIA-Web Face includes 10,575 subjects and 494,414 images.  This dataset is the largest publicly available face image dataset, second only to the private Facebook dataset.  

The proposed deep architecture is described as a succession of two cascaded networks. The first network is constructed from 16 layers of convolution operations on the inputs, intertwined with Rectified Linear Units (ReLU) non-linear operation and five pooling operations. Weights in each convolutional layer form a sequence of \textit{4-d} tensors. Thus, we show all weights of the first network with $W_1$ and weights of the second network with $W_2$. $W_2$ is separated into two groups: $W_{2,1}$ and $W_{2,2}$ represent all weights in the upper and lower branches respectively. 

We modify the AdaMAX algorithm \cite{kingma2014adam} to optimize the network's cost functions. The cost functions represented in (1) and (2) show that how two branches of the second network communicate information and update their learning parameters with each other. AdaMAX optimizer is a variant of Adam optimizer working based on the infinity norm. This version of Adam is a robust and well-adapted optimizer to a verity 
of non-convex optimization problems in the field deep neural networks. All parameter values used in the modified AdaMAX optimizer are initialized based on authors suggestion: $ \alpha = 0.002$, $ \beta_1 = 0.9 $ , $\beta_2=0.999$ and learning rate is set to $ \frac{\alpha}{1-\beta_1(t)} $ showing that learning rate changes over time step. $\mathcal{L}_1$ and $\mathcal{L}_2$ described in (1) and (2) are loss functions designed to perform attribute prediction and face recognition tasks respectively. We use the cross entropy cost function to optimize network parameters.

$T$ and $X=\{x_1,x_2,...,x_i,..x_N\}$ indicate number of facial attributes used in the model and training samples respectively. $L_i'$ and $L_{ji}$ represent identity face label and facial attribute label of number \textit{j} for training sample number \textit{i} respectively. $f$ and $g$ functions are  outputs of the model for attribute prediction and face recognition tasks respectively. As it is shown in (1) and (2) $f$  function (i.e. attribute prediction output) takes the first network parameters (i.e. $W_1$). $g$  function (i.e. face recognition output) takes $W_1$ and $f$ as input.  Thus, both attribute prediction and face recognition use $W_1$ as shared parameters.  Moreover, attribute prediction parameters are used for face recognition. Optimization mainly consists of two steps which first calculates the gradient of the cost functions with respect to the model parameters and then updates the biased first moment estimate, exponentially weighted infinity norm and the model parameters successively.
\begin{equation}
\mathcal{L}_1(W_1,W_{2,1})=\sum\limits_{j=1}^{T} \sum\limits_{i=1}^{N} log(f(x_i,W_1,W_{2,1})\bullet L_{ji}
\end{equation}
\begin{equation}
\begin{split}
\mathcal{L}_2(W_1,W_{2,1},W_{2,2})& =\sum\limits_{i=1}^{N} log(g(x_i,W_1,W_{2,2},\\
&f(x_i,W_1,W_{2,1}))\bullet L'_{i}
\end{split}
\end{equation}
\begin{algorithm}
  \KwData{Face images}
  \KwResult{Model parameters:$\  W_1,W_{2,1},W_{2,2}$}
  initialization\;
  $\alpha \gets \textit{stepsize}$\\
  $\beta_1,\beta_2 \in {[}0,1{)}\textit{\ exponential decay rates}$ \\
  $\theta_1 \gets [W_1,W_{2,1}],\theta_2 \gets [W_1,W_{2,1},W_{2,2}]$\\
  $\mathcal{L}_1(\theta_1),\mathcal{L}_2(\theta_2) \textit{\ stochastic objective functions}$\\
  $\theta_1(0),\theta_2(0) \gets \textit{pretrained model using CASSIA-WF }$\\
  $t \gets   \textit{0\ \   time step}$\\
  $m_0 \gets \textit{0\ \ first moment vector}$\\
  $u_0 \gets \textit{0\ \   exponentially weighted infinity norm}$\\
  \While{$ \theta_1(t) \textit{\ \textbf{and}\ }\theta_2(t)\textit{\ \textbf{not} converged}$}{
  
  $t \gets   t+1$\\
  %\textit{take gradiant w.r.t $\theta_1$ at time step t:}\\
  $g_t(\mathcal{L}_1(\theta_1(t-1))) \gets  \nabla_{\theta_1}\mathcal{L}_1(\theta_1(t-1))$\\
  %\textit{update biased first moment estimates:}\\
  $m_t(\theta_1) \gets \beta_1.m_{t-1}(\theta_1)+(1-\beta_1).g_t(\theta_1)$\\
 %\textit{update the exponentially weighted infinity norm:}\\
  $u_t(\theta_1)\gets max(\beta_2.u_{t-1}(\theta_1),|g_t(\theta_1)|)$\\
  \textbf{update $ W_1,W_{2,1}:$} \\
  $\theta_1(t)\gets \theta_1(t-1)-\frac{\alpha}{1-\beta_1(t)}.\frac{m_t(\theta_1)}{u_t(\theta_1)}$\\  
    $g_t(\mathcal{L}_2(\theta_2(t-1))) \gets  \nabla_{\theta_2}\mathcal{L}_2(\theta_2(t-1))$\\
  %\textit{update biased first moment estimates:}\\
  $m_t(\theta_2) \gets \beta_1.m_{t-1}(\theta_2)+(1-\beta_1).g_t(\theta_2)$\\
 %\textit{update the exponentially weighted infinity norm:}\\
  $u_t(\theta_2)\gets max(\beta_2.u_{t-1}(\theta_2),|g_t(\theta_2)|)$\\
  \textbf{update $ W_1,W_{2,1},W_{2,2}:$} \\
  $\theta_2(t)\gets \theta_2(t-1)-\frac{\alpha}{1-\beta_1(t)}.\frac{m_t(\theta_2)}{u_t(\theta_2)}$\\ 
  }
  \textbf{return $ W_1,W_{2,1},W_{2,2}$ }
  \caption{Modified AdaMax optimizer}
\end{algorithm}
We iterate this algorithm for all the training batch trough several epochs till we convergence in training error. 
\begin{figure*}[ht]
\includegraphics[scale=0.41]{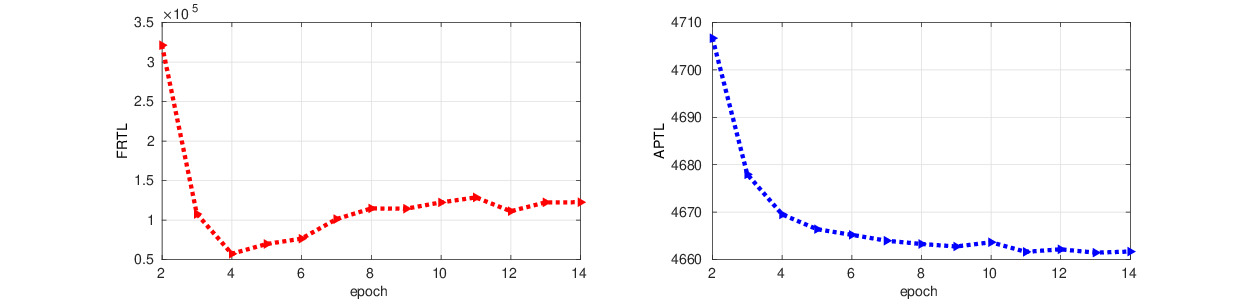}
\begin{center}
\caption{shows Face Recognition\  (\textbf{FRTL})\  and Attribute Prediction\ (\textbf{APTL})\ loss functions minimizing trend on the training data  using \textbf{AAO}.The figure shows that two tasks (i.e. face recognition and attribute Prediction) are trained simultaneously without creating problem on optimizing other loss function.}
\end{center}
\label{fig:X}
\end{figure*}
\section{Experiment}
 We performed experiments in three different cases to examine if privileged data  boost our model performance in recognition and prediction tasks. In first case we train and test the model without using facial attributes while in second and third cases we utilize identity facial attributes concatenated with last pooling layer of $ Net_1 $. Identity facial attributes such as gender, race and face shape are those attributes remaining same in all images from a person.  In second case we use ground truth of face attributes as privileged data in our model. In third case we  predict facial attributes assuming that such information is not  available in test phase; in this case  Softmax outputs of attributes prediction task are concatenated with last pooling layer of $ Net_1 $ for recognizing face images. Experimental results show that  using attribute with  ground truth (i.e. second case ) results in highest boosting in recognition performance and training convergence. Moreover, results show that using attribute with predicted values (i.e. third case) increases recognition performance, however, it decreases training convergence in comparison to the second case.
\subsection{Datasets}
 We tested our model on three face datasets including Celebrity A\cite{liu2015deep}, LFW \cite{huang2007labeled} and Public Figures \cite{kumar2009attribute} in which images are labeled by identities and face attributes.
\subsubsection{Public Figures Face Database}
 Public Figures face dataset is a large, real-world face dataset containing  58,797 images of 200 people collected from Internet using people's name as a search query on variety of image search engines such as Flicker and Google Image. Every image in this dataset is annotated by 73 face attributes such as gender, hair style, age, nationality, noise shape, etc. In this experiment, we select a subset of the Public Figures face dataset containing 8523 face images from 60 different people in the development set. We select 60 identity facial attributes out of 73 attributes describing people without concerning if these attributes change in different images of a person. For example,  race, gender and shape of nose remain same in different images of a person; however, attributes such as wearing glass, mustache, beard  may exist or not  in different images of a person. We discard such attributes in our model because we look for robust as well as discriminative features for describing images in the feature space.
We use   ground truth of facial attributes as it is annotated in \cite{christie2014predicting} ; the annotations are collected from Amazon Mechanical Turk by 10 workers.There are 60 people in development subset of Public Figures; there are on average 270 images per person in this dataset. We select 80\% images of each person as training  and remaining 20\% as test  in our experiment. 
\begin{figure}[h]
\includegraphics[scale=0.23]{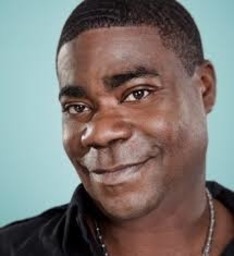}
\includegraphics[scale=0.23]{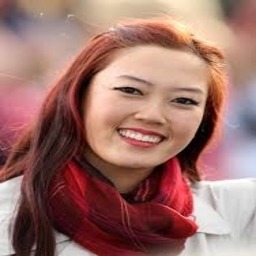}
\includegraphics[scale=0.23]{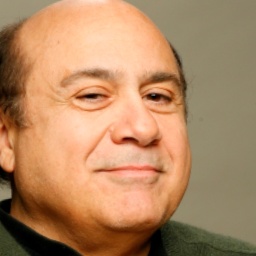}
\includegraphics[scale=0.23]{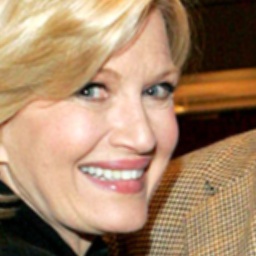}
\indent\begin{tabular}{@{}*{4}{p{.12\textwidth}@{}}}
  \scriptsize Male&\scriptsize Female&\scriptsize Male& \scriptsize Female\\
  \scriptsize Black&\scriptsize Asian& \scriptsize Bald&\scriptsize Blond hair\\ 
\scriptsize Big nose &  \scriptsize Narrow eyes & \scriptsize Round face &  \scriptsize White\\ \\
\end{tabular}
\caption{Some images taken from development set of Public Figure   with corresponding identity attributes;such facial attributes are used as auxiliary data for face recognition.}
\label{fig:pub}
\end{figure}
\begin{table*}[ht]
\centering
\begin{tabular}{c|c|c|c|c|c|c|c|c|c}

  &  Bald  & Black & Black Hair & Blond Hair  & Male & White & Oval Face & Narrow Eyes & Asian \\
\hline
Kumar\cite{kumar2009attribute} & 83.22\%   & 88.65\% & 80.32\% & 78.05\%  & 81.22\% & 91.48\% & 70.26\%&90.02\% &92.32\% \\ 
\hline 
PPA  &  91.25\%  & 86.73\%  &82.05\%  & 79.19\%  & 91.12\% &94.45\%   &72.17\% & 64.97\% &96.73\%  \\
\hline
PPAF   &    93.46\%  & 88.48\%  & 82.27\% & 79.66\%  & 93.98\% &95.66\%   &73.89\% & 66.9\% &98.35\%   \\
\end{tabular}
\caption{Public Figure attribute comparison}\label{eval_table}
\end{table*}\\
We use Softmax-cross entropy as loss function to train the model; Softmax operation is applied on the network outputs to enforce their sum to one; the network outputs represent probability distribution across discrete mutual explicit alternative. After applying this operation, we use cross entropy  to measure distance between  probabilities obtained by the model and ground truth. We use modified AdaMax Optimizer (\textbf{AMO}) to minimize cost functions; Figure. 2 shows \textbf{AMO} minimizes face recognition and attribute prediction cost functions on training data; it turns out that networks cost functions designed for face recognition and attribute prediction are optimized simultaneously without creating problem for each other training loss convergence.
Experimental results show that facial attributes contribute to increasing face recognition performance; to verify this claim, we performed experiments in three different cases described earlier in Experiment section. In the third case we emphasize on predicting facial attributes  because in real and typical face recognition problem, such information is not available in test step. To use facial attributes information as an auxiliary modality in the model for face recognition, we fuse this modality with the last pooling layer of the model shown in Figure 1. Depending on the cases described earlier; this fusion is either ground truth or  predicted values obtained in the model. Table 3  shows that  using attribute with ground truth (i.e. first case) results in the best performance in terms of accuracy. Moreover, using predicted attributes in the model (i.e. third case) outperforms the first case  which does not use any privilege data; however, this improvement in recognition performance results in decreasing training convergence rate; in the other word, it takes more epochs to train the model fully. Figure. 3 show that accuracy of training in the beginning is less than the case disregarding facial attributes.
\begin{figure}
\includegraphics[trim={0cm 0 0 0},scale=0.5]{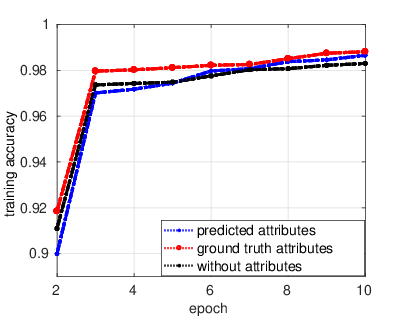}
\begin{center}
\caption{comparing training convergence between three scenarios }
\label{fig:conv1}
\end{center}
\end{figure}
\begin{figure}[h]
\includegraphics[trim={0cm 0 0 0},scale=0.18]{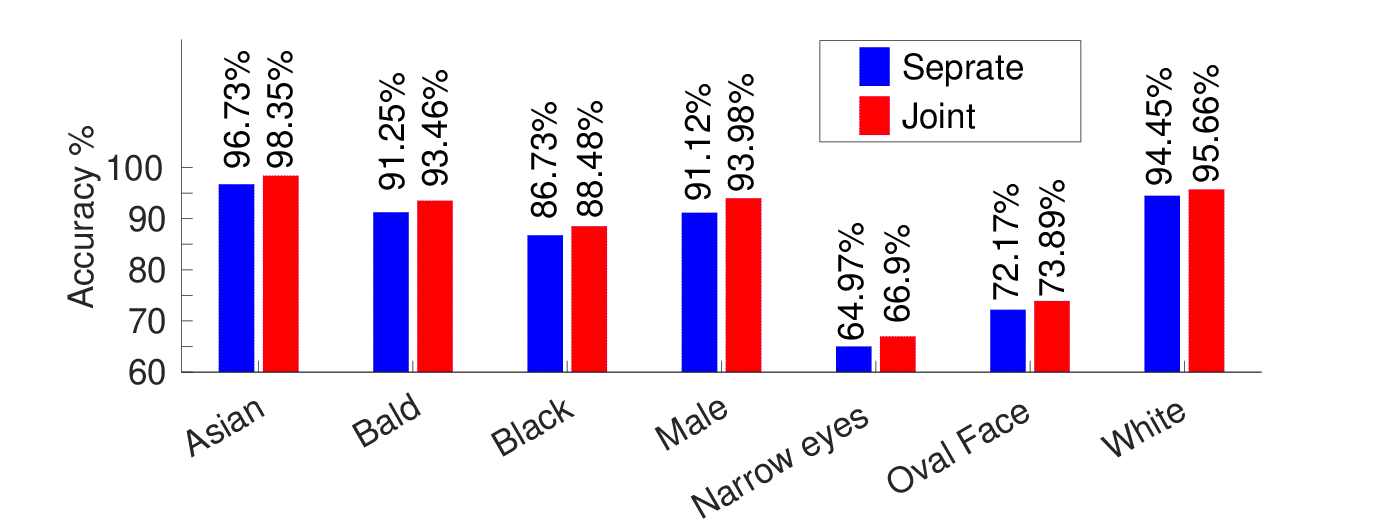}
\begin{center}
\caption{Comparing facial identity attributes prediction performance trained separately and jointly with face recognition task; Figure 5 shows that attribute prediction performance increases if the model is trained jointly by face recognition task.}
\label{fig:com1}
\end{center}
\end{figure}

Experimental results show that training two tasks jointly increases both face recognition  as well as facial attribute classification performance specially  on identity face attributes such as  gender and nationality. For example, experiments performed on the Public Figures indicate that performance on identity attributes including male, Asian, white, black , bald, narrow eyes and oval face classification is improved around 2\% on average if the tasks are trained jointly. Moreover, experiment (Table 1) turns out the model outperforms accuracy reported in \cite{kumar2009attribute} for identity facial attributes classification. Figure. 4 show the model performance on attribute prediction in two cases which attributes are trained separately from face and the case which attributes are trained jointly with face. Results show that the model trained jointly classifies identity attributes better than the case which the model is trained separately. One of the intuitive reasons causing this improvement is once a deep CNN model is trained to recognize face images; it  learns face attributes as well indirectly. In other words these two modalities enhance each others' performance once they are trained jointly.

We finally report performance of the model for face recognition in three cases on the test data; Table 1 indicates that using facial attributes as privileged data  boosts the model performance on face recognition task.
\begin{table*}[h]
\centering
\begin{tabular}{c|c|c|c}
  & First case  & Second case & Third case \\
\hline
 Public Figure & 95.89\%   & 98.43\% & 97.38\% \\ 
\hline 
 LFW & 97.10\%  & 99.73\% & 99.66\% \\
\hline
CelebA &  96.30\% & 98.87\% & 97.99\% \\
\end{tabular}
\caption{Comparing attribute prediction models on LFW dataset.}\label{eval_table}
\end{table*}
\subsubsection{Labeled Face in the wild}
 \begin{figure*}[th!]
\includegraphics[trim={0cm 0 0 0},scale=0.4]{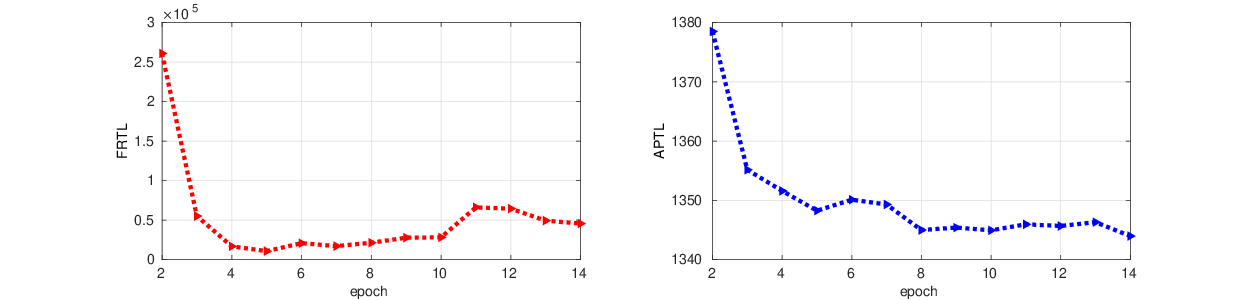}
\begin{center}
\caption{shows Face Recognition\  (\textbf{FRTL})\  and Attribute Prediction\ (\textbf{APTL})\ loss functions minimizing trend on the training data  using \textbf{AAO}.The figure shows that two tasks (i.e. face recognition and attribute Prediction) are trained simultaneously without creating problem on optimizing other loss function.}
\end{center}
\end{figure*}
LWF is a very well-known for face recognition as well as attribute classification. This dataset fills gap existing in unconstrained condition for face recognition problem.
This dataset contains more than 13,000 images of faces collected from Internet. Each face has been labeled with the name of the people. We use 21 identity attributes to train the model; we select 80\% of images as training data and 20\% as test data.
\begin{figure}[h]
\includegraphics[scale=0.23]{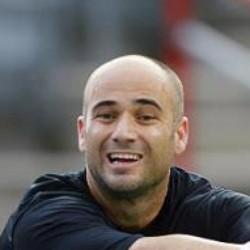}
\includegraphics[scale=0.23]{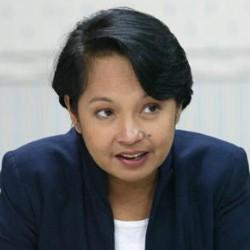}
\includegraphics[scale=0.23]{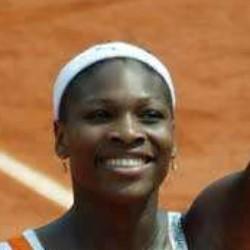}
\includegraphics[scale=0.23]{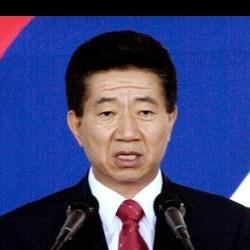}
\noindent\begin{tabular}{@{}*{4}{p{.13\textwidth}@{}}}
  \scriptsize Male&\scriptsize Female&\scriptsize Female& \scriptsize Male\\
  \scriptsize Bald&\scriptsize Black \scriptsize hair&\scriptsize Big nose& \scriptsize Narrow eyes\\
\scriptsize Young &  \scriptsize Straight hair & \scriptsize Black &  \scriptsize Big nose\\
\end{tabular}
\caption{Some images with corresponding attributes taken from development subset of PubFig database; }
\end{figure}
\begin{figure}[h]
\includegraphics[trim={0cm 0 0 0},scale=0.58]{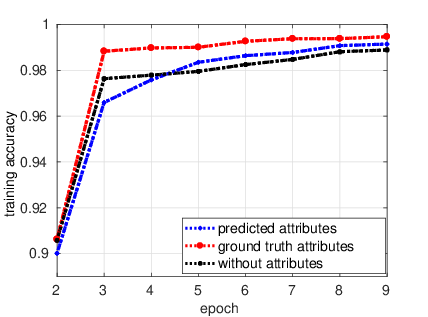}
\begin{center}
\caption{comparing training convergence between three scenarios }
\end{center}
\end{figure}
\begin{figure}[h!]
\includegraphics[trim={0cm 0 0 0},scale=0.18]{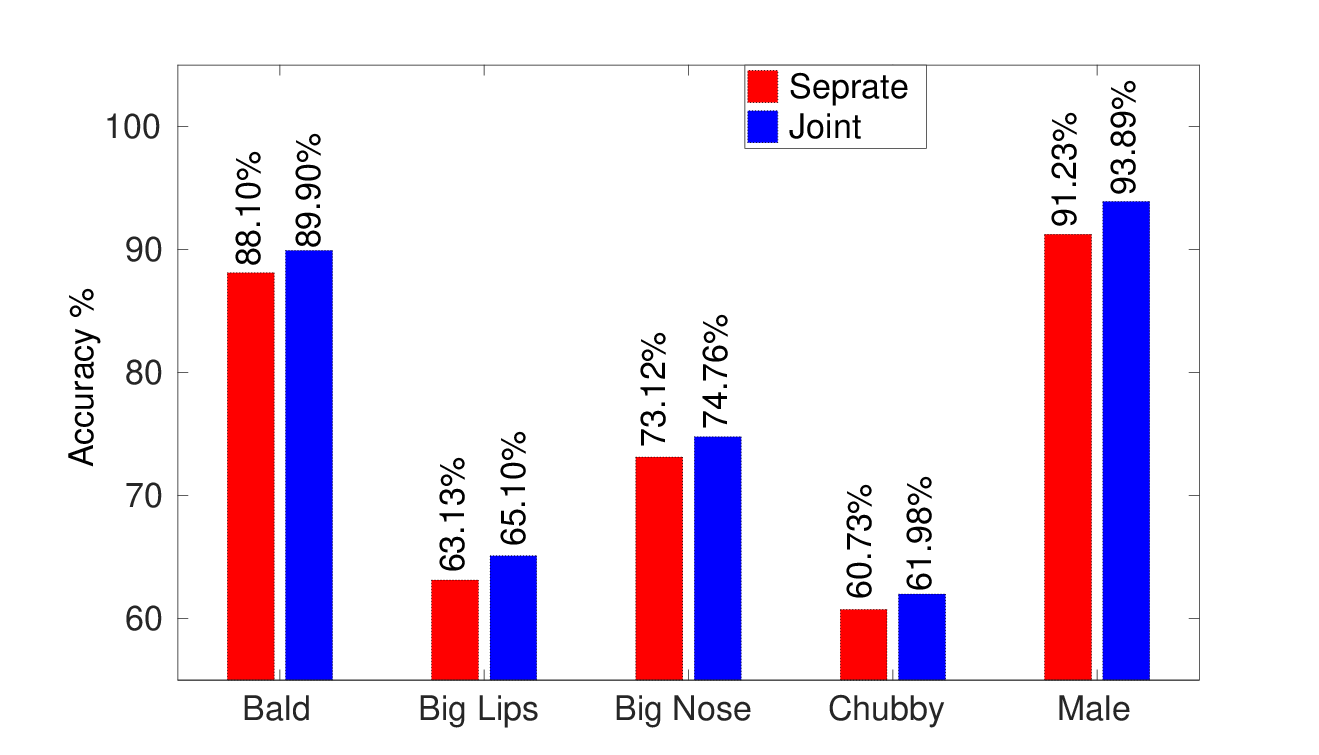}
\begin{center}
\caption{Comparing facial identity attributes prediction performance trained separately and jointly with face recognition task; Figure 5 shows that attribute prediction performance increases if the model is trained jointly by face recognition task.}
\end{center}
\end{figure}
we compare our model with some other current and well-performed models including Deep Face \cite{taigman2014deepface}, Face Net \cite{schroff2015facenet}, Deep FR \cite{parkhi2015deep}, Baidu \cite{liu2015targeting} and Center Face \cite{wen2016discriminative} having high accuracy on LFW dataset; the results show the superiority of our model in comparison to these models.
We also compare our model with some current models such as FaceTracer \cite{kumar2008facetracer} PANDA \cite{zhang2014panda} and state of the art LNets+ANet \cite{liu2015deep}  attribute prediction models on CelebA and LFW datasets. Results present that our model is comparable with these models specially in identity facial attributes such as gender, hair color, nose and lips shape, chubby and bald prediction. Results indicate that the model classify identity facial attributes as good as state of the art method and still is comparable to other well performed methods; the model in some cases specially for gender classification outperforms the best method.

%%\begin{figure}[h!]
%%\includegraphics[trim={0cm 0 0 0},scale=0.155]{LFWComparing1.eps}
%%\begin{center}
%%\caption{Face recognition performance on test data in three scenarios; GT, PA and NPD represent performance of the model when it uses ground truth, predicted attributes and no privilege data respectively.   }
%%\end{center}
%%\end{figure}
\begin{figure}[h!]
\includegraphics[trim={0cm 0 0 0},scale=0.17]{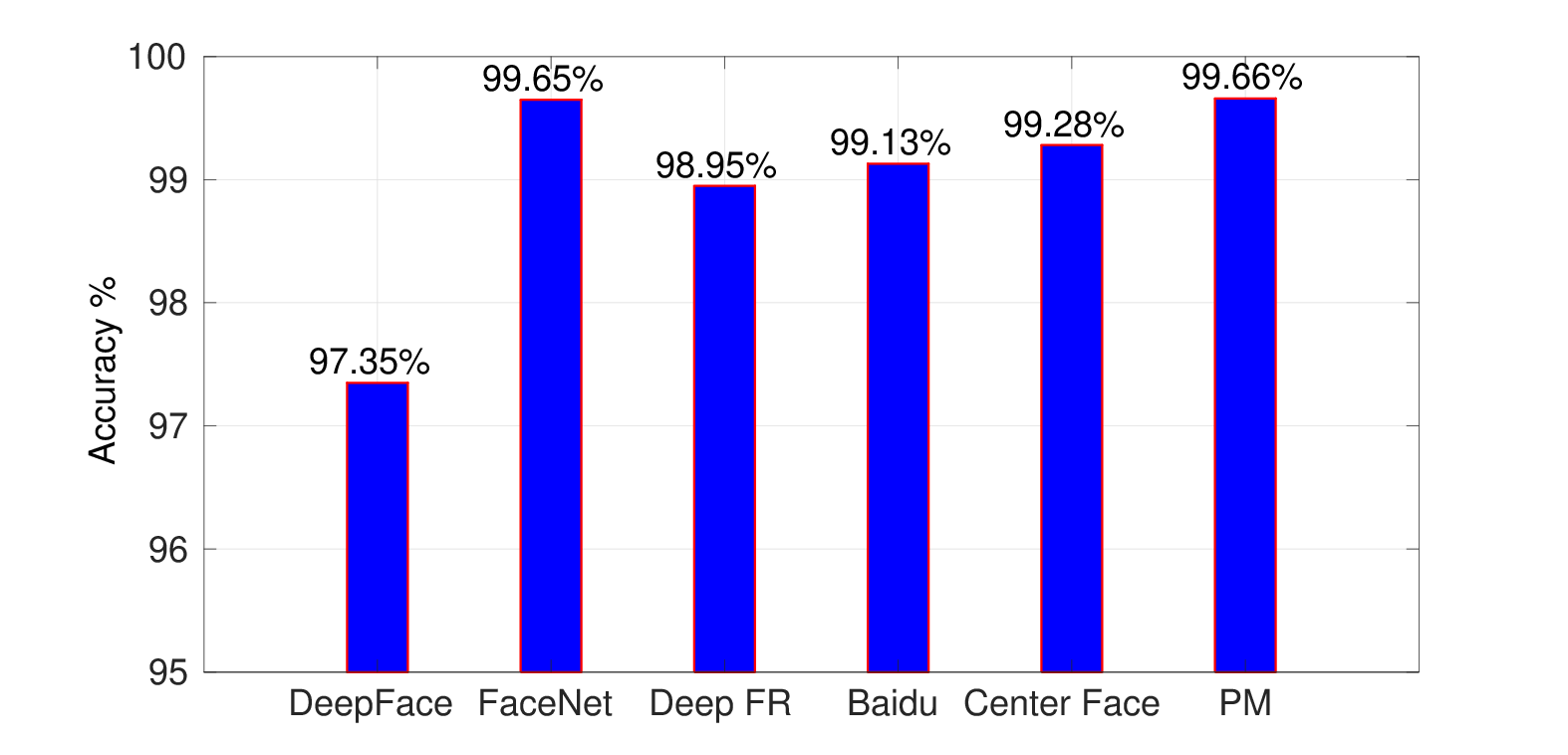}
\begin{center}
\caption{Face recognition performance on test data in three scenarios; GT, PA and NPD represent performance of the model when it uses ground truth, predicted attributes and no privilege data respectively.   }
\end{center}
\end{figure}
\begin{table*}[ht]
\centering
\begin{tabular}{c|c|c|c|c|c|c|c}
  &  Bald  & Big Lips & Big Nose & Chubby  & Male & Black Hair  & Blond Hair \\
\hline
 FaceTracer & 77\%   & 68\% & 73\% & 67\%  & 84\% &76\% & 88\%\\ 
%\hline 
 %PANDA-w &  82\%  & 64\% & 71\%  & 65\%  &  86\% & 78\%& 87\%\\
\hline
 PANDA &  84\% & 73\% & 79\% & 69\%  & 92\% & 87\% & 94\% \\
 %\hline
% LNets+ANet(w/o) &  83\% & 72\% & 76\% & 70\%  & 91\% & 86\% &94\%\\
 \hline
 LNets+ANet &  88\%  & 75\% & 81\% & 73\%  & 94\% & 90\% &97\%\\
 \hline
 Our model & 90\%   & 65\% & 75\% & 62\%  & 94\% & 80\% & 88\%
\end{tabular}
\caption{Comparing attribute prediction models on LFW dataset.}\label{eval_table}
\end{table*}
\begin{table*}[ht]
\centering
\begin{tabular}{c|c|c|c|c|c|c|c}
  &  Bald  & Big Lips & Big Nose & Chubby  & Male & Black Hair  & Blond Hair \\
\hline
 FaceTracer & 89\%   & 64\% & 74\% & 86\%  & 91\% &70\% & 80\%\\ 
%\hline 
 %PANDA-w &  92\%  & 61\% & 70\%  & 82\%  &  93\% & 74\%& 81\%\\
\hline
 PANDA &  96\% & 67\% & 75\% & 86\%  & 97\% & 85\% & 93\% \\
% \hline
 %LNets+ANet(w/o) &  95\% & 66\% & 75\% & 86\%  & 94\% & 84\% &91\%\\
 \hline
 LNets+ANet &  98\%  & 68\% & 78\% & 91\%  & 98\% & 88\% &95\%\\
 \hline
 Our model & 97\%   & 67\% & 79\% & 91\%  & 99\% & 75\% & 84\%
\end{tabular}
\caption{Comparing attribute prediction models on CelebFacesA dataset.}\label{eval_table}
\end{table*}
\subsubsection{CelebFaces Attributes Dataset }
CelebA is a large-scale face attributes and rich annotated dataset containing more than 200K celebrity images, each of which is notated with  40 attributes. CelebA has about ten thousand identities with  twenty images per identity on average. In this experiment, we select 20 identities having more than 30 images. we select 21 identity facial attributes such as chubby and gender out of 40 attributes. We select 80 \% images of each person as training  and rest as test data in our experiment. 
\begin{figure}[h]
\includegraphics[scale=0.32]{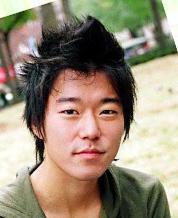}
\includegraphics[scale=0.32]{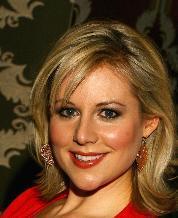}
\includegraphics[scale=0.32]{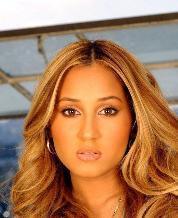}
\includegraphics[scale=0.32]{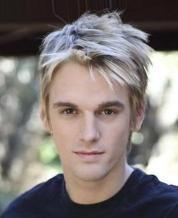}
\indent\begin{tabular}{@{}*{4}{p{.13\textwidth}@{}}}
 \scriptsize Male&\scriptsize Female&\scriptsize Female& \scriptsize Male\\
 \scriptsize Narrow eyes&\scriptsize Blond hair& \scriptsize Oval face&\scriptsize Straig1ht hair \\
\scriptsize Black hair &  \scriptsize High cheekbones & \scriptsize Pale skin &  \scriptsize Young\\
\end{tabular}
\caption{Some images with corresponding attributes taken from development subset of PubFig}
\end{figure}
\begin{figure}[h!]
\includegraphics[trim={0cm 0 0 0},scale=0.174]{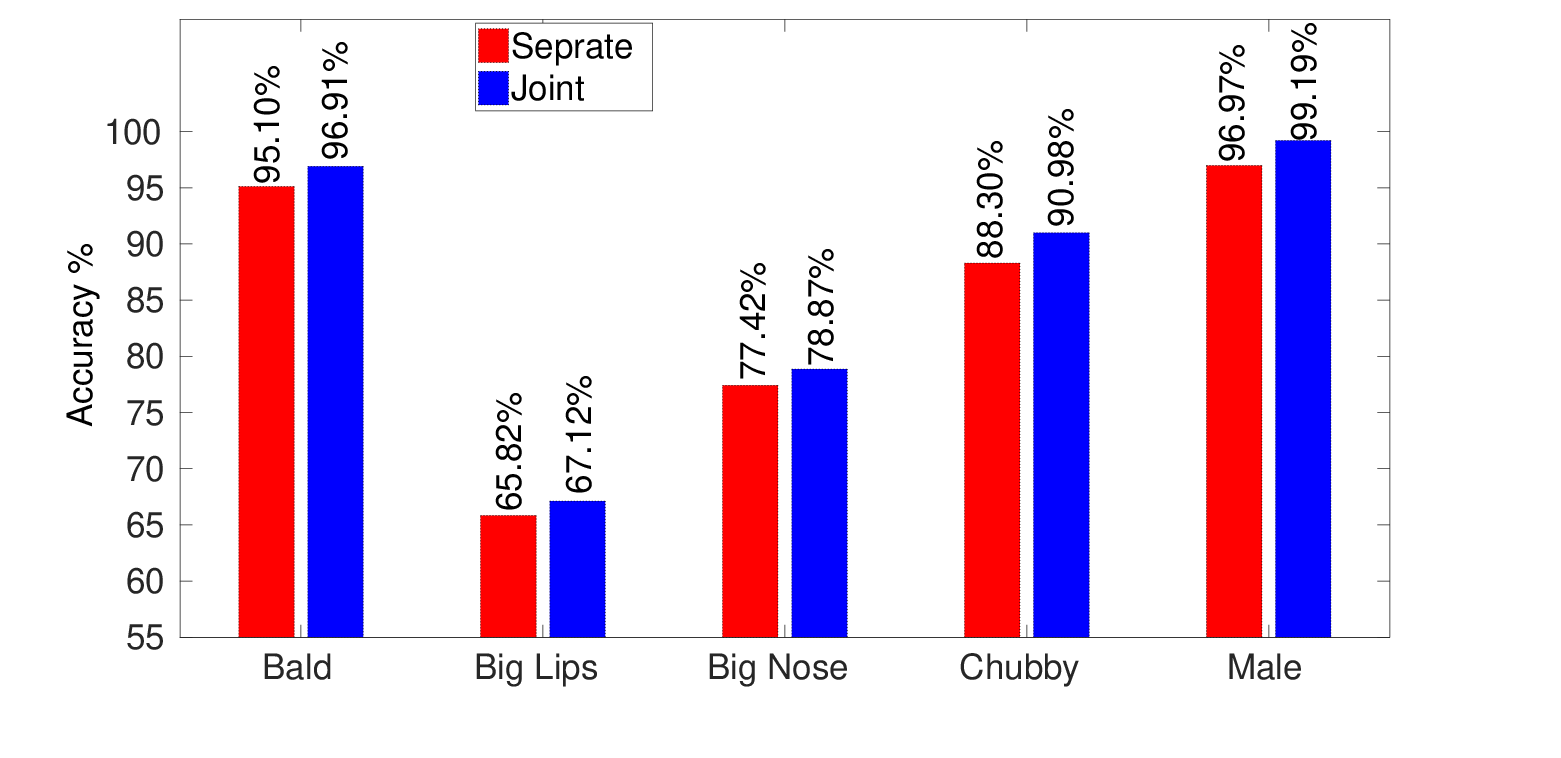}
\begin{center}
\caption{Comparing facial identity attributes prediction performance trained separately and jointly with face recognition task; Figure 5 shows that attribute prediction performance increases if the model is trained jointly by face recognition task.}
\end{center}
\end{figure}

\subsubsection{Conclusion}
In this paper, we proposed a model predicting facial attributes to use as privileged modality for increasing face recognition performance. The model predicts attributes as well as recognizes face images jointly through shared learning parameters. The model increases both face recognition and face attribute prediction performance in comparison to the case which the model is trained separately. Experimental results indicate superiority of the model in comparison to state-of-the-art face recognition models; the model classifies identity facial attributes as good as state of the art. The model improves face recognition performance without concerning if privilege data is available or not in test step due to predicting them in the model.
{\small
\bibliographystyle{ieee}
\bibliography{egbib}

\begin{thebibliography}{10}\itemsep=-1pt

\bibitem{akata2013label}
Z.~Akata, F.~Perronnin, Z.~Harchaoui, and C.~Schmid.
\newblock Label-embedding for attribute-based classification.
\newblock In {\em Proceedings of the IEEE Conference on Computer Vision and
  Pattern Recognition}, pages 819--826, 2013.

\bibitem{berg2013poof}
T.~Berg and P.~N. Belhumeur.
\newblock Poof: Part-based one-vs.-one features for fine-grained
  categorization, face verification, and attribute estimation.
\newblock In {\em Proceedings of the IEEE Conference on Computer Vision and
  Pattern Recognition}, pages 955--962, 2013.

\bibitem{bourdev2011describing}
L.~Bourdev, S.~Maji, and J.~Malik.
\newblock Describing people: A poselet-based approach to attribute
  classification.
\newblock In {\em Computer Vision (ICCV), 2011 IEEE International Conference
  on}, pages 1543--1550. IEEE, 2011.

\bibitem{bourdev2016pose}
L.~D. Bourdev.
\newblock Pose-aligned networks for deep attribute modeling, July~26 2016.
\newblock US Patent 9,400,925.

\bibitem{Cao_THN_2017}
Z.~Cao, M.~Long, J.~Wang, and Q.~Yang.
\newblock Transitive hashing network for heterogeneous multimedia retrieval.
\newblock 2017.

\bibitem{christie2014predicting}
G.~Christie, A.~Parkash, U.~Krothapalli, and D.~Parikh.
\newblock Predicting user annoyance using visual attributes.
\newblock In {\em Proceedings of the IEEE Conference on Computer Vision and
  Pattern Recognition}, pages 3630--3637, 2014.

\bibitem{chung2012deep}
J.~Chung, D.~Lee, Y.~Seo, and C.~D. Yoo.
\newblock Deep attribute networks.
\newblock In {\em Deep Learning and Unsupervised Feature Learning NIPS
  Workshop}, volume~3, 2012.

\bibitem{gong2013hidden}
D.~Gong, Z.~Li, D.~Lin, J.~Liu, and X.~Tang.
\newblock Hidden factor analysis for age invariant face recognition.
\newblock In {\em Proceedings of the IEEE International Conference on Computer
  Vision}, pages 2872--2879, 2013.

\bibitem{goodfellow2014generative}
I.~Goodfellow, J.~Pouget-Abadie, M.~Mirza, B.~Xu, D.~Warde-Farley, S.~Ozair,
  A.~Courville, and Y.~Bengio.
\newblock Generative adversarial nets.
\newblock In {\em Advances in neural information processing systems}, pages
  2672--2680, 2014.

\bibitem{hemulti}
Y.~He, L.~Chen, and J.~Chen.
\newblock Multi-task relative attributes prediction by incorporating local
  context and global style information features.

\bibitem{huang2007labeled}
G.~B. Huang, M.~Ramesh, T.~Berg, and E.~Learned-Miller.
\newblock Labeled faces in the wild: A database for studying face recognition
  in unconstrained environments.
\newblock Technical report, Technical Report 07-49, University of
  Massachusetts, Amherst, 2007.

\bibitem{iscen2019label}
A.~Iscen, G.~Tolias, Y.~Avrithis, and O.~Chum.
\newblock Label propagation for deep semi-supervised learning.
\newblock In {\em Proceedings of the IEEE Conference on Computer Vision and
  Pattern Recognition}, pages 5070--5079, 2019.

\bibitem{izadi2020optimization}
M.~R. Izadi, Y.~Fang, R.~Stevenson, and L.~Lin.
\newblock Optimization of graph neural networks with natural gradient descent.
\newblock {\em arXiv preprint arXiv:2008.09624}, 2020.

\bibitem{kazemi2018unsupervised2}
H.~Kazemi, F.~Taherkhani, and N.~M. Nasrabadi.
\newblock Unsupervised facial geometry learning for sketch to photo synthesis.
\newblock In {\em 2018 International Conference of the Biometrics Special
  Interest Group (BIOSIG)}, pages 1--5. IEEE, 2018.

\bibitem{kingma2014adam}
D.~Kingma and J.~Ba.
\newblock Adam: A method for stochastic optimization.
\newblock {\em arXiv preprint arXiv:1412.6980}, 2014.

\bibitem{krizhevsky2012imagenet}
A.~Krizhevsky, I.~Sutskever, and G.~E. Hinton.
\newblock Imagenet classification with deep convolutional neural networks.
\newblock In {\em Advances in neural information processing systems}, pages
  1097--1105, 2012.

\bibitem{kumar2008facetracer}
N.~Kumar, P.~Belhumeur, and S.~Nayar.
\newblock Facetracer: A search engine for large collections of images with
  faces.
\newblock In {\em European conference on computer vision}, pages 340--353.
  Springer, 2008.

\bibitem{kumar2009attribute}
N.~Kumar, A.~C. Berg, P.~N. Belhumeur, and S.~K. Nayar.
\newblock Attribute and simile classifiers for face verification.
\newblock In {\em Computer Vision, 2009 IEEE 12th International Conference on},
  pages 365--372. IEEE, 2009.

\bibitem{liu2015targeting}
J.~Liu, Y.~Deng, T.~Bai, Z.~Wei, and C.~Huang.
\newblock Targeting ultimate accuracy: Face recognition via deep embedding.
\newblock {\em arXiv preprint arXiv:1506.07310}, 2015.

\bibitem{liu2019exploiting}
X.~Liu, J.~Van De~Weijer, and A.~D. Bagdanov.
\newblock Exploiting unlabeled data in cnns by self-supervised learning to
  rank.
\newblock {\em IEEE transactions on pattern analysis and machine intelligence},
  2019.

\bibitem{liu2015deep}
Z.~Liu, P.~Luo, X.~Wang, and X.~Tang.
\newblock Deep learning face attributes in the wild.
\newblock In {\em Proceedings of the IEEE International Conference on Computer
  Vision}, pages 3730--3738, 2015.

\bibitem{luo2013deep}
P.~Luo, X.~Wang, and X.~Tang.
\newblock A deep sum-product architecture for robust facial attributes
  analysis.
\newblock In {\em Proceedings of the IEEE International Conference on Computer
  Vision}, pages 2864--2871, 2013.

\bibitem{mohsenvand2020contrastive}
M.~N. Mohsenvand, M.~R. Izadi, and P.~Maes.
\newblock Contrastive representation learning for electroencephalogram
  classification.
\newblock In {\em Machine Learning for Health}, pages 238--253. PMLR, 2020.

\bibitem{parkhi2015deep}
O.~M. Parkhi, A.~Vedaldi, A.~Zisserman, et~al.
\newblock Deep face recognition.
\newblock In {\em BMVC}, volume~1, page~6, 2015.

\bibitem{schroff2015facenet}
F.~Schroff, D.~Kalenichenko, and J.~Philbin.
\newblock Facenet: A unified embedding for face recognition and clustering.
\newblock In {\em Proceedings of the IEEE Conference on Computer Vision and
  Pattern Recognition}, pages 815--823, 2015.

\bibitem{taherkhani2017restoring}
F.~Taherkhani and M.~Jamzad.
\newblock Restoring highly corrupted images by impulse noise using radial basis
  functions interpolation.
\newblock {\em IET Image Processing}, 12(1):20--30, 2017.

\bibitem{taherkhani2019matrix}
F.~Taherkhani, H.~Kazemi, and N.~M. Nasrabadi.
\newblock Matrix completion for graph-based deep semi-supervised learning.
\newblock In {\em Thirty-Third AAAI Conference on Artificial Intelligence},
  2019.

\bibitem{taherkhani2018deep}
F.~Taherkhani, N.~M. Nasrabadi, and J.~Dawson.
\newblock A deep face identification network enhanced by facial attributes
  prediction.
\newblock In {\em Proceedings of the IEEE conference on computer vision and
  pattern recognition workshops}, pages 553--560, 2018.

\bibitem{taherkhani2018facial}
F.~Taherkhani, V.~Talreja, H.~Kazemi, and N.~Nasrabadi.
\newblock Facial attribute guided deep cross-modal hashing for face image
  retrieval.
\newblock In {\em 2018 International Conference of the Biometrics Special
  Interest Group (BIOSIG)}, pages 1--6. IEEE, 2018.

\bibitem{taigman2014deepface}
Y.~Taigman, M.~Yang, M.~Ranzato, and L.~Wolf.
\newblock Deepface: Closing the gap to human-level performance in face
  verification.
\newblock In {\em Proceedings of the IEEE conference on computer vision and
  pattern recognition}, pages 1701--1708, 2014.

\bibitem{talreja2018using}
V.~Talreja, F.~Taherkhani, M.~C. Valenti, and N.~M. Nasrabadi.
\newblock Using deep cross modal hashing and error correcting codes for
  improving the efficiency of attribute guided facial image retrieval.
\newblock In {\em 2018 IEEE Global Conference on Signal and Information
  Processing (GlobalSIP)}, pages 564--568. IEEE, 2018.

\bibitem{talreja2019attribute}
V.~Talreja, F.~Taherkhani, M.~C. Valenti, and N.~M. Nasrabadi.
\newblock Attribute-guided coupled gan for cross-resolution face recognition.
\newblock {\em arXiv preprint arXiv:1908.01790}, 2019.

\bibitem{torfason2016face}
R.~Torfason, E.~Agustsson, R.~Rothe, and R.~Timofte.
\newblock From face images and attributes to attributes.
\newblock In {\em Asian Conference on Computer Vision}, pages 313--329.
  Springer, 2016.

\bibitem{wen2016latent}
Y.~Wen, Z.~Li, and Y.~Qiao.
\newblock Latent factor guided convolutional neural networks for age-invariant
  face recognition.
\newblock In {\em Proceedings of the IEEE Conference on Computer Vision and
  Pattern Recognition}, pages 4893--4901, 2016.

\bibitem{wen2016discriminative}
Y.~Wen, K.~Zhang, Z.~Li, and Y.~Qiao.
\newblock A discriminative feature learning approach for deep face recognition.
\newblock In {\em European Conference on Computer Vision}, pages 499--515.
  Springer, 2016.

\bibitem{Yang_2017_Pairwise}
E.~Yang, C.~Deng, W.~Liu, X.~Liu, D.~Tao, and X.~Gao.
\newblock Pairwise relationship guided deep hashing for cross-modal retrieval.
\newblock In {\em Proc. AAAI Conference on Artificial Intelligence}, Feb. 2017.

\bibitem{zhang2014panda}
N.~Zhang, M.~Paluri, M.~Ranzato, T.~Darrell, and L.~Bourdev.
\newblock Panda: Pose aligned networks for deep attribute modeling.
\newblock In {\em Proceedings of the IEEE conference on computer vision and
  pattern recognition}, pages 1637--1644, 2014.

\bibitem{zhong2016face}
Y.~Zhong, J.~Sullivan, and H.~Li.
\newblock Face attribute prediction using off-the-shelf cnn features.
\newblock In {\em Biometrics (ICB), 2016 International Conference on}, pages
  1--7. IEEE, 2016.

\end{thebibliography}
}
\end{document}